\pdfoutput=1

\documentclass[11pt]{article}

\usepackage[final]{acl}

\usepackage{times}
\usepackage{latexsym}

\usepackage[T1]{fontenc}

\usepackage[utf8]{inputenc}

\usepackage{microtype}

\usepackage{inconsolata}

\usepackage{graphicx}
\usepackage{epstopdf, epsfig}
\usepackage{xcolor}
\usepackage{mdframed}
\usepackage{enumitem}
\usepackage[utf8]{inputenc}
\usepackage[english]{babel}
\usepackage{hyperref}
\usepackage{amsmath}
\usepackage{subcaption}
\usepackage{float}
\usepackage{booktabs}
\usepackage{wrapfig}
\usepackage{float} 
\usepackage{placeins}
\usepackage{graphicx}
\usepackage{array}
\usepackage{tcolorbox}
\usepackage{caption}
\usepackage{multirow}
\usepackage[strict]{changepage}

\definecolor{customgreen}{HTML}{009900}
\definecolor{customred}{HTML}{FF0000}

\newmdenv[
  backgroundcolor=gray!10,
  linewidth=0pt,
  innerleftmargin=10pt,
  innerrightmargin=10pt,
  innertopmargin=10pt,
  innerbottommargin=10pt
]{codeblock}

\title{The Multicultural Medical Assistant: Can LLMs Improve
Medical ASR Errors Across Borders?}

\author{Ayo Adedeji \textsuperscript{1} \,
  Mardhiyah Sanni \textsuperscript{2} \,
  Emmanuel Ayodele \textsuperscript{2} \,
  Sarita Joshi \textsuperscript{1} \,
  Tobi Olatunji \textsuperscript{2} \\
  \\
  \textsuperscript{1} Google Cloud, 1600 Amphitheatre Parkway, Mountain View, CA 94043, USA \\
  \textsuperscript{2} Intron Health, San Francisco, CA, USA and Lagos, Nigeria
}

\begin{document}
\maketitle

\begin{abstract}
The global adoption of Large Language Models (LLMs) in healthcare shows promise to enhance clinical workflows and improve patient outcomes. However, Automatic Speech Recognition (ASR) errors in critical medical terms remain a significant challenge. These errors can compromise patient care and safety if not detected. This study investigates the prevalence and impact of ASR errors in medical transcription in Nigeria, the United Kingdom, and the United States. By evaluating raw and LLM-corrected transcriptions of accented English in these regions, we assess the potential and limitations of LLMs to address challenges related to accents and medical terminology in ASR. Our findings highlight significant disparities in ASR accuracy across regions and identify specific conditions under which LLM corrections are most effective.
\end{abstract}

\vspace{0.01em}
\section{Introduction}\label{sec:introduction}
In recent years, medical Automatic Speech Recognition (ASR) systems have become integral to healthcare, revolutionizing processes such as physician-dictated notes, telemedicine, and doctor-patient conversations \citep{Johnson2014, Zhang2023}. By easing the administrative burden on healthcare providers, these systems allow them to focus more on patient care and less on documentation \citep{Saxena2018}.

Despite their contributions to efficiency in modern healthcare recordkeeping, significant challenges remain. Achieving high accuracy across various medical terminologies and diverse demographic accents continues to be a formidable task \citep{mani-etal-2020-towards, DiChristofano2023}. ASR systems often struggle with the precise recognition of specialized medical terminology, including drug names and diagnoses \citep{Hodgson2015}. This limitation can lead to errors that undermine the quality and reliability of medical records.

Recent advances in Large Language Models (LLMs) have emerged as a promising direction for addressing these challenges. Their ability to understand and process human language nuances positions them as potential tools to improve the accuracy of medical transcription. Our study evaluates how effectively LLMs can improve medical transcription accuracy across healthcare settings in Nigeria, the United Kingdom, and the United States.

\vspace{0.2em}
Our \textbf{contributions} are:

1. The first large-scale evaluation of both ASR performance and LLM-based corrections across healthcare settings in Nigeria, the United Kingdom, and the United States, analyzing 191 medical conversations spanning multiple specialties.

2. Our evaluation framework and metrics, released for reproducibility and future research in cross-regional medical ASR.

\section{Related Work}\label{sec:related-work}
\subsection{ASR in Medical Settings}
Recent research has focused on improving the accuracy of ASR to drive its acceptance in clinical practice. Although ASR systems have shown promise in dictating medical reports, they often struggle with the nuanced context and speaker diarization inherent in patient-clinician conversations. The high word error rate (WER) in these scenarios indicates poor performance in contextual understanding and diarization \citep{Park2023, Tran2023}. The wide variability in accents between healthcare providers and patients exacerbates these issues, leading to possible misinterpretations of critical medical information \citep{afonja2024performantasrmodelsmedical, Zaporowski2024}.

Addressing these challenges requires extensive training data, yet privacy concerns and regulatory restrictions have led to a scarcity of publicly available medical conversation datasets \citep{papadopoulos-korfiatis-etal-2022-primock57, le-duc-2024-vietmed}. This limited data availability has restricted researchers' ability to develop robust and adaptable ASR systems for global healthcare environments.

\subsection{Error Correction Approaches}
Recent advances in error correction methods have shown promising results. Leng et al. \citep{leng-etal-2021-fastcorrect-2} introduced FastCorrect 2, a non-autoregressive model that takes advantage of multiple ASR candidates to improve correction accuracy through a novel alignment algorithm. Boros et al. \citep{boros-etal-2024-post} evaluated fourteen foundation LLMs for post-transcription correction, providing valuable information on the capabilities and limitations of LLMs in this domain. Complementing these approaches, Radhakrishnan et al. \citep{radhakrishnan-etal-2023-whispering} developed a cross-modal fusion technique that combines acoustic information with external linguistic representations, demonstrating significant improvements in WER reduction.

\subsection{LLMs in Medical Transcription}
Initial experiments with GPT-4 to create structured documentation from clinical conversations revealed consistent challenges in information preservation, often introducing errors through omission or addition of content \citep{kernberg2024using}. However, more promising results have emerged from LLM-enhanced ASR systems, which have demonstrated improved speaker diarization and reduced WER. \citep{adedeji2024sound, wang2024diarizationlmspeakerdiarizationpostprocessing}.

\subsection{Cross-Regional ASR Studies}
While several studies have examined ASR performance in specific healthcare contexts, comprehensive cross-regional evaluations remain limited. Existing research has focused mainly on single-region or single-accent scenarios, leaving a significant gap in our understanding of the performance of ASR systems in global healthcare settings \citep{DiChristofano2023}. Our work addresses this gap by providing a systematic evaluation of both ASR performance and LLM-based corrections across Nigeria, the United Kingdom, and the United States, offering insight into the practical applicability of ASR and LLM-correction approaches in global healthcare settings.

\section{Materials}\label{sec:materials}

\begin{table}[h!]
\centering
{\footnotesize
\renewcommand{\arraystretch}{1.2}
\begin{tabular}{|m{1.8cm}|m{1.7cm}|m{1.2cm}|m{1.5cm}|}
\hline
\textbf{Dataset} & \textbf{Region} & \textbf{Num. Conv.} & \textbf{Avg. Turns} \\
\hline
Intron Health Teleconsulta-tions & Africa & 25 & 99 \\
\hline
PriMock57 & UK / Europe & 57 & 92 \\
\hline
Fareez Medical Interviews & United States & 109 & 112 \\
\hline
\end{tabular}
\caption{Overview of the three medical conversation datasets used in this study, showing the geographic distribution, number of conversations, and average number of turns per conversation.}
\label{tab:conversation_summary}
}
\end{table}

\subsection{Nigerian Dataset} 
For our Nigerian dataset, we used a collection of simulated medical conversations provided by Intron Health \citep{olatunji2023afrispeech}. This dataset comprises 25 doctor-patient consultations that capture approximately 4 hours of spoken dialogue.

These consultations span multiple specialties, with obstetrics and gynecology being the most prevalent (5 cases), followed by infectious diseases (4), gastroenterology (2), cardiology (2), and endocrinology (2). Additional specialties include neurology (2), orthopedics (2), general surgery (2), and single cases in pulmonology, otolaryngology, family medicine, and hematology, providing comprehensive coverage across medical disciplines.

Each consultation in this dataset is rich in content, averaging 99 conversational turns and 1,437 spoken words. To ensure demographic representation, the dataset features a mix of 4 female and 7 male speakers, all aged between 25 and 35 years. Notably, both the patient and the doctor roles are portrayed by Nigerian medical professionals.

\subsection{United Kingdom Dataset}
For our United Kingdom dataset, we used PriMock57, a collection of simulated medical conversations developed by Babylon Health \citep{papadopoulos-korfiatis-etal-2022-primock57}. This dataset comprises 57 doctor-patient consultations that capture approximately 9 hours of spoken dialogue.

These consultations span multiple conditions, with cardiovascular problems being the most prevalent (11 cases), followed by gastrointestinal disorders (8), respiratory conditions (8), migraines (6), and other infections (8). Additional categories include fever (4), dermatological conditions (4), anaphylactic reactions (3), mental health concerns (3), and physical injuries (2), providing comprehensive coverage of primary care presentations.

To ensure demographic diversity, the dataset includes 7 clinicians and 57 actors portraying patients, with a balanced gender distribution and an age range predominantly between 25 to 45 years. The linguistic landscape of the UK's healthcare system is reflected in the variety of accents: clinicians speak mainly British English with some Indian accents, while patient accents span British English (47.4\%), various European (31.6\%), and other English and non-English accents (21\%).

\subsection{United States Dataset}
For our United States dataset, we used a subset of simulated medical conversations developed by Fareez et al. \citep{Fareez2022}. This subset comprises 109 doctor-patient consultations that capture approximately 22 hours of spoken dialogue, strategically sampled from a larger pool of 272 conversations. 

To ensure specialty diversity, we included all non-respiratory cases and randomly sampled from the respiratory cases, which formed the majority of the original dataset. These consultations span multiple conditions, with respiratory cases being the most prevalent (51 cases), followed by musculoskeletal conditions (46), gastrointestinal (6), cardiac (5), and dermatologic (1). 

The dataset features a balanced gender distribution between doctors (57\% male, 43\% female) and patients (55\% male, 45\% female). Both doctor and patient roles were portrayed by senior medical students and resident doctors, leveraging their clinical experience to simulate realistic patient-doctor interactions.

\section{Approach}

Our approach combines ASR systems and LLMs to transcribe, diarize, correct, and analyze medical conversations. We employ six ASR systems, chosen for their capabilities in handling medical terminology and diverse accents, and three LLMs, selected for their reasoning capabilities. For a detailed description of these models, see Appendices \ref{subsec:appendix-asr-models} and \ref{subsec:appendix-llms}.

To identify and categorize medical concepts within our transcriptions, we utilized Google Cloud's Healthcare Natural Language API \citep{GoogleCloudHealthcareNLP}. This API allows for the precise extraction and categorization of medical entities, which is crucial for our analysis. More details about this API can be found in Appendix \ref{subsec:appendix-medical-concepts}.

\subsection{Pipeline}
\begin{figure}[t]
\centering
\includegraphics[width=0.38\columnwidth]{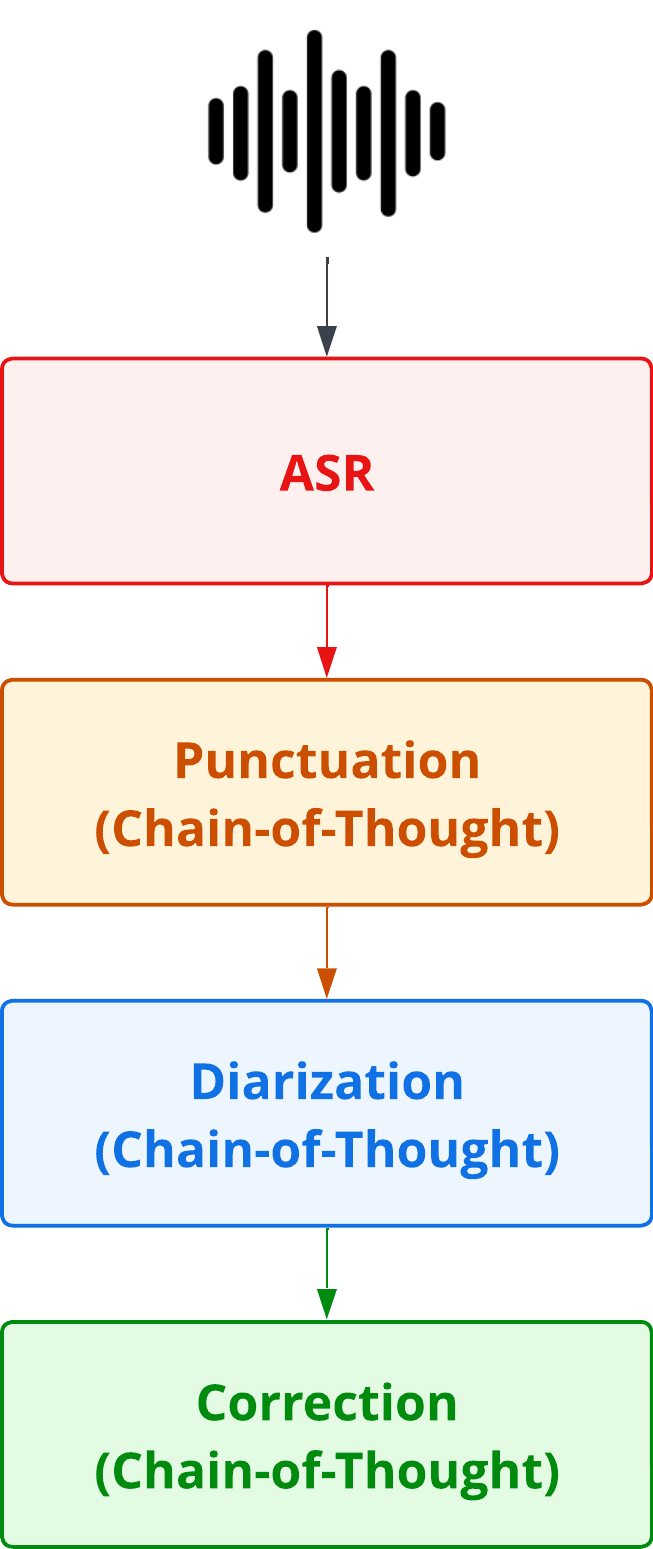}
\caption{The steps of our Chain-of-Thought (CoT) prompting pipeline for medical conversation processing.}
\label{fig:pipeline}
\end{figure}

Our pipeline for processing and analyzing medical conversations builds upon the methodology established in our previous work \citep{adedeji2024sound}, with key enhancements to address cross-regional accents and variations in English. Figure \ref{fig:pipeline} illustrates the main stages of our approach.

\vspace{0.2em}
The pipeline consists of the following stages:

\begin{enumerate}[leftmargin=1em, itemsep=0pt, topsep=0pt]
  \item \textbf{Transcription:} Raw audio from our datasets is processed by each ASR system to generate initial transcriptions.

  \item \textbf{Punctuation Enhancement:} We instruct the LLM with a Chain-of-Thought (CoT) prompt to improve punctuation to normalize orthographic variations and provide structured outputs for subsequent steps. A representative prompt can be found in Appendix \ref{sec:appendix-example-punctuation-prompt}.

  \item \textbf{Diarization:} We instruct the LLM with a CoT prompt to perform speaker diarization from scratch, without relying on any existing speaker labels. The LLM analyzes conversation patterns to assign each word and sentence to distinct speakers. A representative prompt can be found in Appendix \ref{sec:appendix-example-diarization-prompt}.
  
  \item \textbf{Correction:} We instruct the LLM with a CoT prompt to improve transcription accuracy while preserving medical context. A representative prompt can be found in Appendix \ref{sec:appendix-example-correction-prompt}.

  \item \textbf{Error Analysis and Entity Recognition:} We use Google Cloud's Healthcare Natural Language API for medical entity identification and categorization and the Jiwer Python library \citep{Jiwer} for WER computation. This allows us to calculate both standard WER for general transcription accuracy and Medical Concept Word Error Rate (MC-WER) for evaluating medical concept transcription accuracy.
\end{enumerate}

\vspace{1em}
\subsection{Evaluation Metrics}
To quantify the effectiveness of our pipeline, we employ metrics that capture three key aspects of performance: transcription accuracy, diarization performance, and preservation of medical concepts.

\subsubsection{Transcription Accuracy}
We used WER as our primary metric to assess overall transcription accuracy. WER is calculated as:
\begin{equation}
    WER = \frac{S + D + I}{N}
\end{equation}
where $S$ is the number of substituted words, $D$ is the number of deleted words, $I$ is the number of inserted words, and $N$ is the total number of words in the reference transcript.

\subsubsection{Diarization Performance}
We adapted a word-level error rate metric, common in multi-party conversation analysis \citep{yu2022comparativestudyspeakerattributedautomatic, shafey2019jointspeechrecognitionspeaker}, to jointly evaluate speaker diarization and transcription accuracy. While traditional diarization error rate (DER) measures speaker attribution accuracy as:
\begin{equation}
\small
DER = \frac{\text{Incorrectly attributed words}}{\text{Total words}} \times 100%
\end{equation}
our analysis uses WER computed against aligned hypothesis and ground truth speaker segments, effectively capturing both speaker attribution and transcription errors. We computed this speaker-level WER separately for doctor and patient speech segments, where a higher WER reflects both poor speaker attribution and transcription errors. This approach eliminates the need for timestamp alignment while providing granular insights into system performance across speaker roles.

\subsubsection{Medical Concept Accuracy}
\label{subsec:Medical Concept Accuracy Metrics}
We used MC-WER to specifically evaluate the transcription accuracy of medical terminology. While standard WER treats all words independently, MC-WER operates on complete medical concepts identified through Healthcare NLP annotation. MC-WER is calculated as:

\begin{equation}
MC\text{-}WER = \frac{S_m + D_m + I_m}{M}
\end{equation}
where $S_m$ is the number of medical concept substitutions (e.g., "diuretics" for "Dioralyte" or "high tension" for "hypertension"), $D_m$ is the number of medical concept deletions, $I_m$ is the number of medical concept insertions, and $M$ is the total number of medical concepts in the reference transcript. Medical concepts are aligned between hypothesis and reference transcripts as complete units, regardless of whether they comprise single or multiple words.
We calculated MC-WER for both lemmatized and non-lemmatized versions of medical concepts:

\begin{itemize}
\item \textbf{Lemmatized MC-WER:} Captures errors in the base forms of medical terms (e.g., treating "antibiotics" and "antibiotic" as equivalent).

\item \textbf{Non-lemmatized MC-WER:} Preserves morphological variations (e.g., distinguishing between singular/plural forms), which can be clinically significant.
\end{itemize}
\vspace{0.5em}
The results presented in this paper use the non-lemmatized version to maintain the integrity of important medical distinctions.

\vspace{1em}
\section{Experiments}

We conducted all Whisper Large V3 and NVIDIA Canary-1B inference experiments using a single NVIDIA T4 GPU. For both models, the transcriptions were processed in 20 second speech slices to optimize memory usage while maintaining transcription accuracy. For the remaining ASR systems, we utilized their publicly available enterprise APIs with default configurations optimized for medical speech. Similarly, all LLM operations were performed through public-facing APIs, with temperature values optimized for each model within the 0 to 0.1 range based on preliminary testing. The LLM correction and diarization steps were applied in 10-line segments. No additional hyperparameter tuning or specific resource configurations were needed for these tasks.

\section{Results}

\subsection{Cross-Dataset WER Analysis}
\FloatBarrier

\begin{figure}[t]
\centering
\includegraphics[width=0.95\columnwidth]{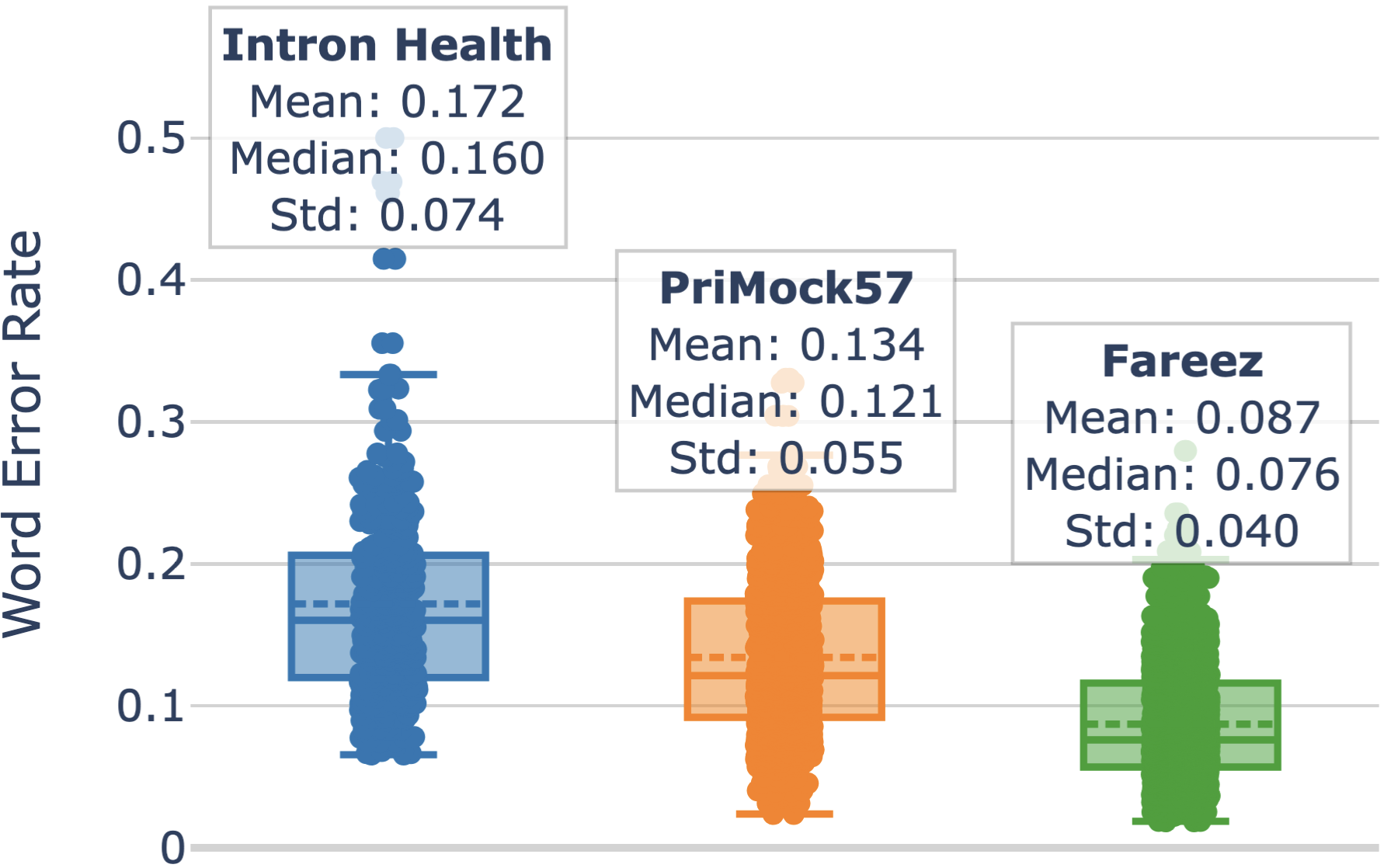}
\caption{
Distribution of WER across baseline ASR transcriptions for all speakers in each dataset.
}

\label{fig:wer-all}
\end{figure}
\begin{figure}[t]
\centering
\includegraphics[width=0.95\columnwidth]{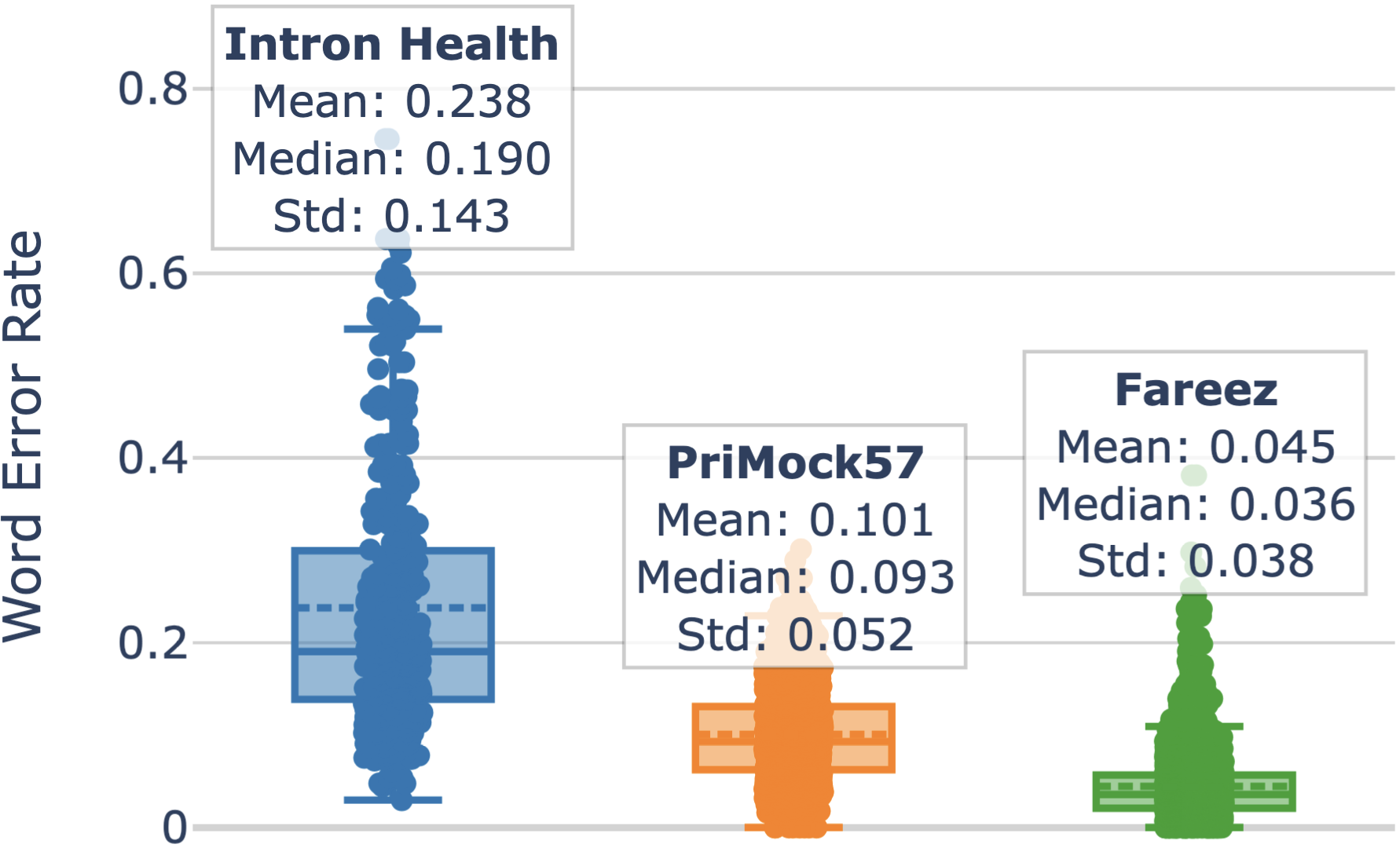}
\caption{
Distribution of MC-WER across baseline ASR transcriptions for all speakers in each dataset.
}
\label{fig:mc-wer-all}
\end{figure}

Our analysis revealed systematic differences in transcription accuracy across the datasets (Figures \ref{fig:wer-all} and \ref{fig:mc-wer-all}). For general transcription accuracy (WER), the Nigerian-accented Intron Health dataset showed the highest error rate (mean: 0.172), followed by the UK-accented PriMock57 (mean: 0.134), and the US-accented Fareez dataset (mean: 0.087). This represents a 4.7 percentage point difference between PriMock57 and Fareez and an 8.5 percentage point difference between Intron Health and Fareez. Medical concept transcription accuracy (MC-WER) showed larger disparities: Intron Health (mean: 0.238), PriMock57 (mean: 0.101), and Fareez (mean: 0.045). This represents a 5.6 percentage point difference between PriMock57 and Fareez and a 19.3 percentage point difference between Intron Health and Fareez.

\subsection{LLM Diarization Performance}

\begin{figure}[t]
    \centering
    \begin{subfigure}{0.95\columnwidth}
        \centering
        \includegraphics[width=\textwidth]{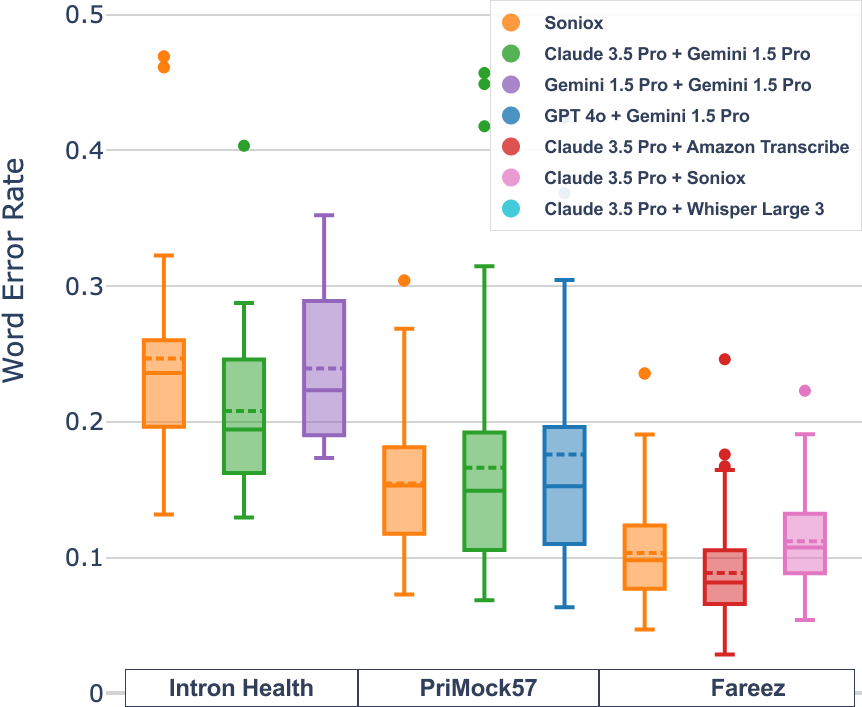}
        \caption{Patient Speech WER Distribution}
        \label{fig:wer-patients}
    \end{subfigure}

    \vspace{1.5em}

    \begin{subfigure}{0.95\columnwidth}
        \centering
        \includegraphics[width=\textwidth]{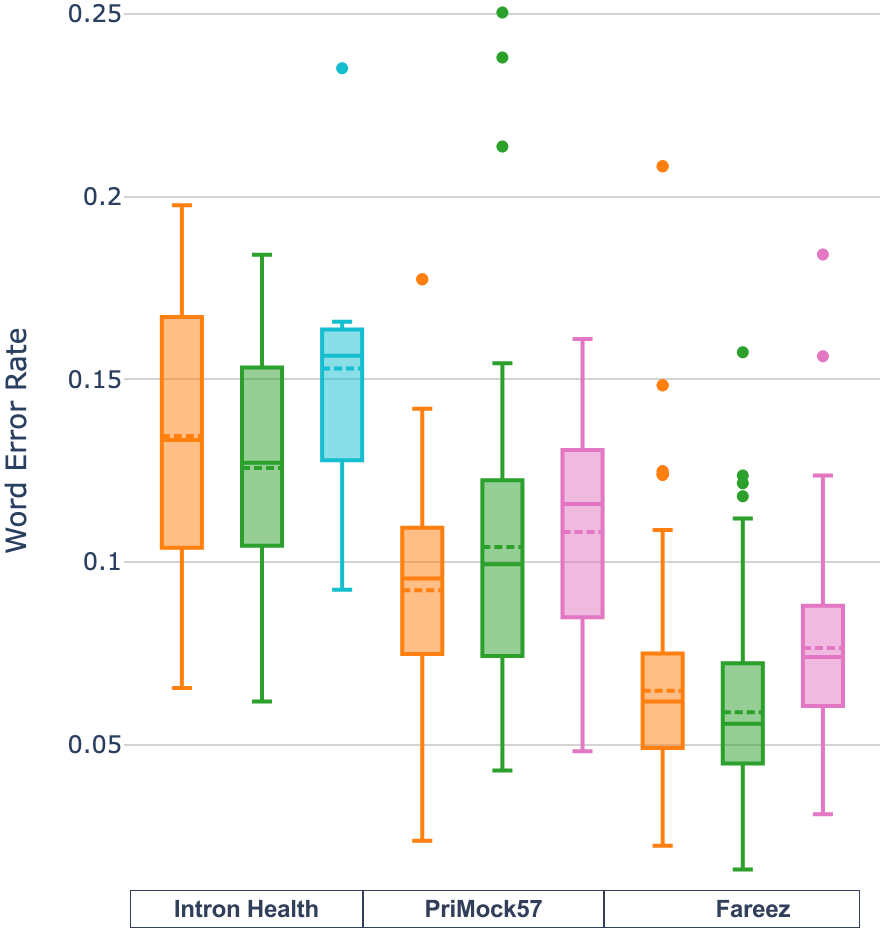}
        \caption{Doctor Speech WER Distribution}
        \label{fig:wer-doctors}
    \end{subfigure}

    \caption{WER Distributions for patient and doctor speech for the baseline ASR system, Soniox (orange), and top performing LLM and ASR pairs.}
\end{figure}

Across the three datasets, LLM-based diarization (where an LLM performs de novo speaker labeling of an ASR-generated transcript) demonstrated varying degrees of improvement when compared to Soniox, our baseline ASR system chosen for its native diarization capabilities. For patient speech (Figure \ref{fig:wer-patients}), the Intron Health dataset showed the most marked improvement, with Claude 3.5 Sonnet + Gemini 1.5 Pro (mean: 0.208) outperforming Soniox (mean: 0.247) by 3.9 percentage points. 

Multiple LLM-ASR pairs achieved comparable performance in PriMock57, with the best pair yielding mean WER of 0.15, a modest 0.5 percentage point improvement over Soniox (mean: 0.155). The Fareez dataset maintained the lowest overall error rates, with the top LLM-ASR pair achieving mean WER of 0.089 versus Soniox's 0.1034, an improvement of 1.44 percentage points.

Doctor speech (Figure \ref{fig:wer-doctors}) demonstrated consistently lower error rates than patient speech across all datasets. For doctor speech, the top performing LLM and ASR pair (Claude 3.5 Sonnet + Gemini 1.5 Pro) achieved mean WERs of 0.127, 0.104, and 0.059 for Intron Health, PriMock57, and Fareez datasets, respectively, compared to Soniox's baseline mean WERs of 0.135, 0.095, and 0.064. The performance differences between LLM-based diarization and Soniox varied by dataset, with small improvements in Intron Health and Fareez, but slightly worse performance in PriMock57 for doctor speech.

\subsection{LLM Correction Performance}

\begin{figure}[t]
    \centering
    \includegraphics[width=0.97\columnwidth]{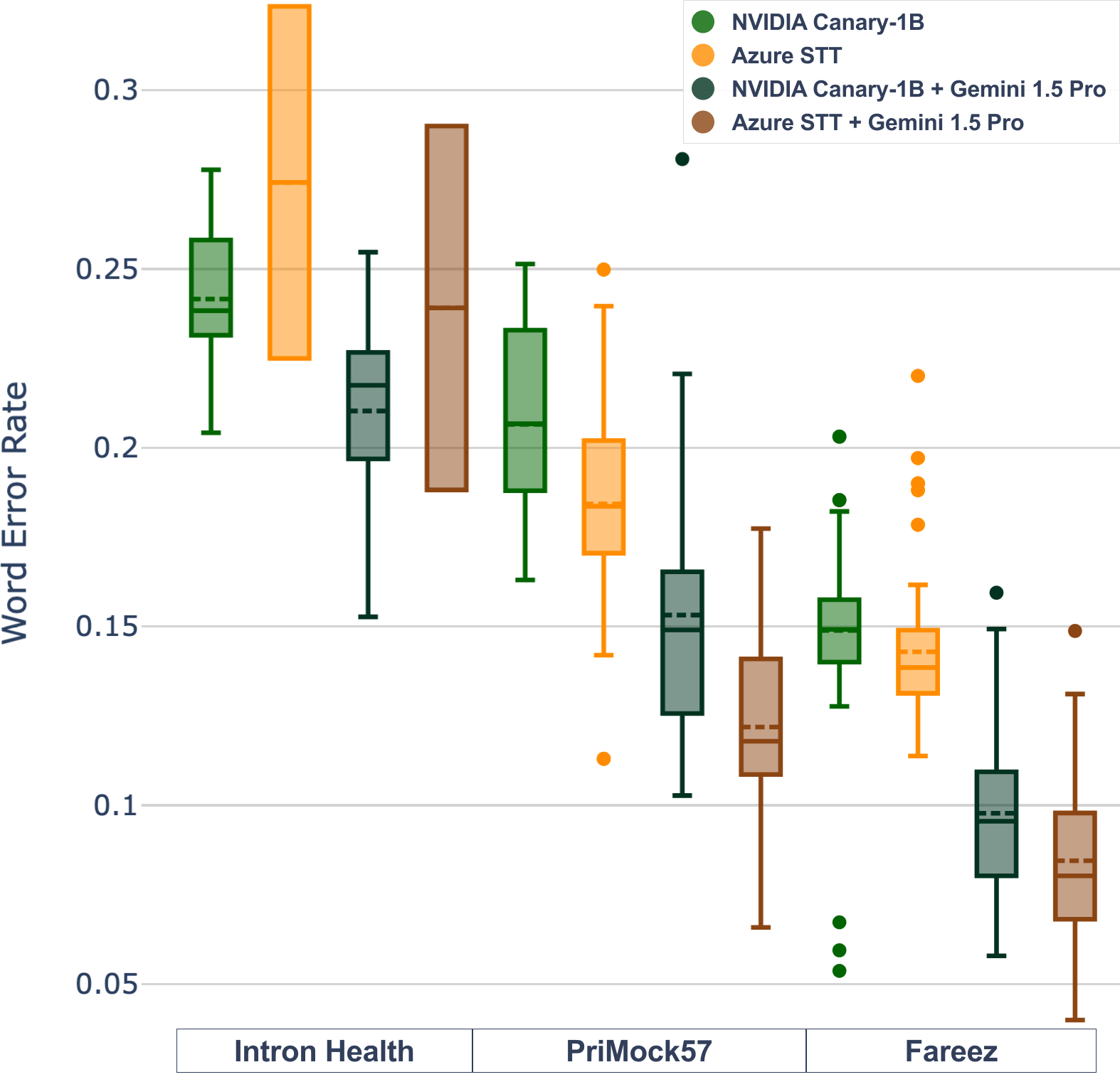}
    \caption{Comparison of WER before and after correction for lower-performing ASR systems across datasets. The graph shows the WER distribution of NVIDIA Canary-1B (green), Azure STT (orange), and their LLM-corrected versions using Gemini 1.5 Pro (olive for NVIDIA correction, brown for Azure STT correction).}
    \label{fig:wer-correction-lower-performing}
\end{figure}

\begin{figure}[t]
    \centering
    \includegraphics[width=0.97\columnwidth]{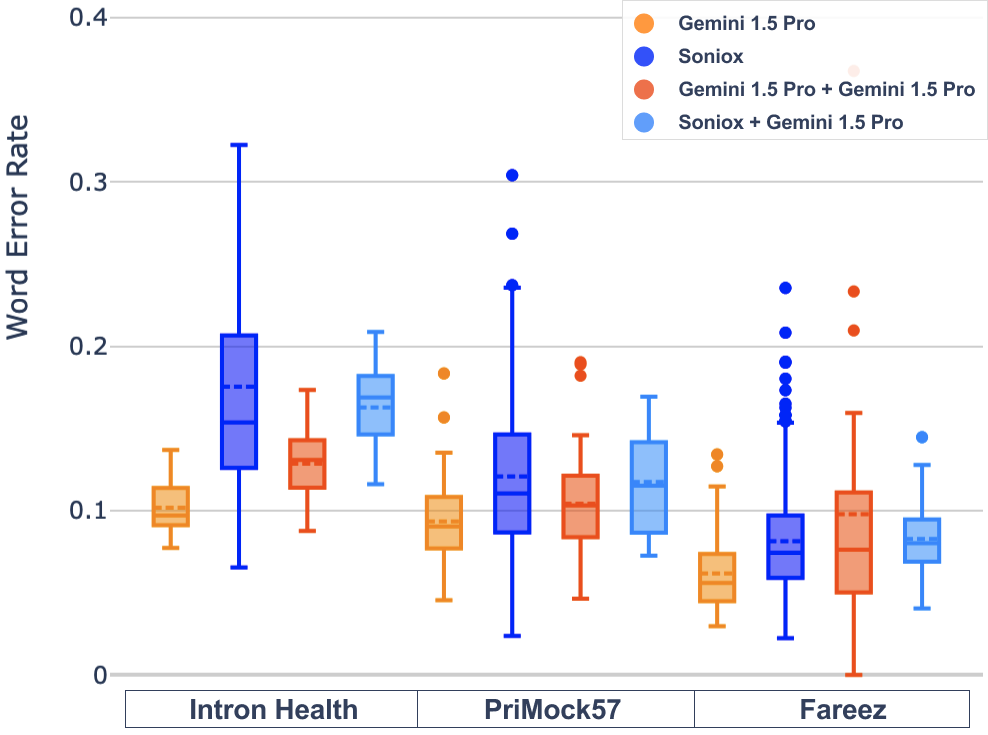}
    \caption{Comparison of WER before and after correction for higher performing ASR systems across datasets. The graph shows the WER distribution of Gemini 1.5 Pro (orange), Soniox ASR (dark blue), and their LLM-corrected versions using Gemini 1.5 Pro (red for Gemini correction, light blue for Soniox correction).}
    \label{fig:wer-correction-higher-performing}
\end{figure}

When evaluating LLM corrections with Gemini 1.5 Pro as our representative model, we observed universal improvements across lower-performing ASR systems (Figure \ref{fig:wer-correction-lower-performing}). Correction results for other LLMs are available in Appendices \ref{subsec:intron health wer table}, \ref{subsec:primock57 wer table}, and \ref{subsec:fareez wer table}. In the Intron Health dataset, NVIDIA Canary-1B's mean WER improved from 0.241 to 0.21, and Azure STT's from 0.274 to 0.239. On PriMock57, corrections reduced Canary-1B's mean WER from 0.207 to 0.153 and Azure STT's from 0.184 to 0.122. The Fareez dataset showed the strongest relative improvements, with Canary-1B's mean WER decreasing from 0.149 to 0.098 and Azure STT's from 0.142 to 0.085.

For higher performing systems such as Gemini 1.5 Pro and Soniox, corrections yielded minimal improvements or negative effects (Figure \ref{fig:wer-correction-higher-performing}). Testing on Intron Health showed Gemini 1.5 Pro's mean WER increased from 0.102 to 0.129, while Soniox's decreased from 0.175 to 0.163. The pattern continued with PriMock57, where Gemini 1.5 Pro's mean WER rose from 0.093 to 0.104 while Soniox's slightly improved from 0.12 to 0.117. On Fareez, both systems showed reduced performance, with Gemini 1.5 Pro's mean WER increasing from 0.062 to 0.098 and Soniox's from 0.081 to 0.083.

\subsection{Error Types and Correction Patterns}

\begin{figure}[t]
    \centering
    \includegraphics[width=1\columnwidth]{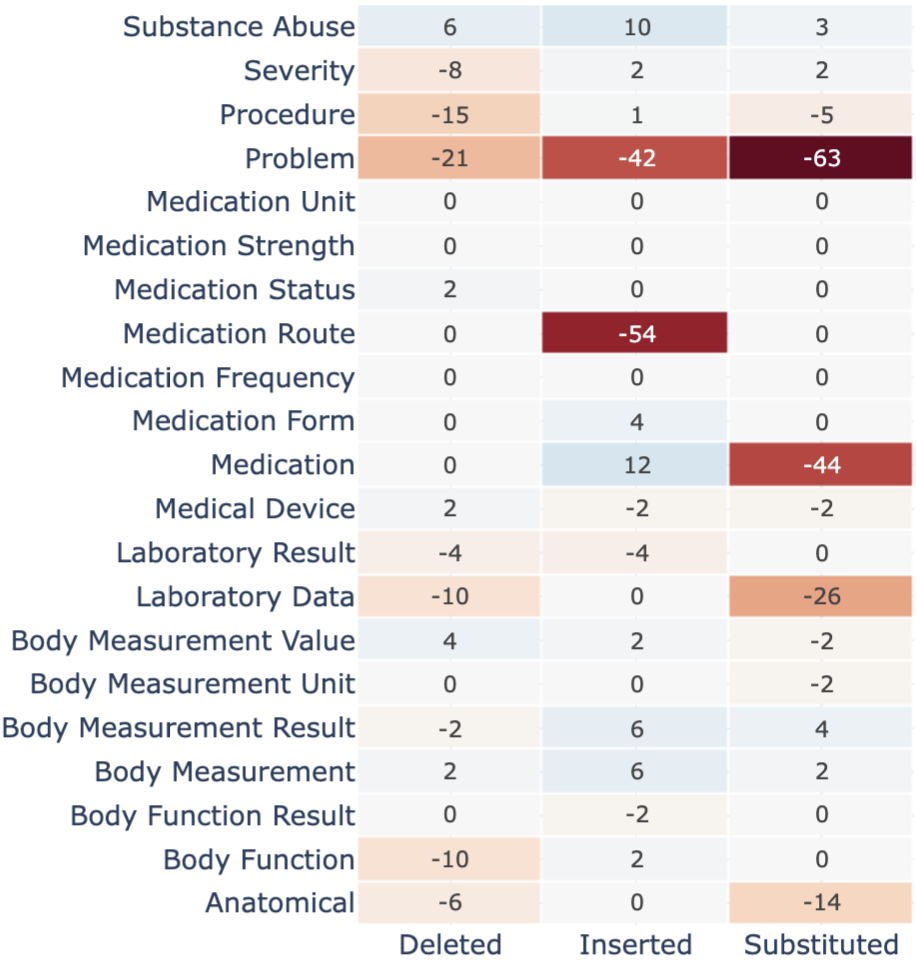}
    \caption{Distribution of error differences by medical concept category, comparing pre-correction (NVIDIA Canary-1B ASR baseline) with post-correction (with Gemini 1.5 Pro) on the Intron Health dataset. Negative values indicate fewer errors after correction, while positive values indicate more errors after correction.}
    \label{fig:nvidia-gemini-char-error-distribution}
\end{figure}

\begin{figure}[t]
    \centering
    \includegraphics[width=1\columnwidth]{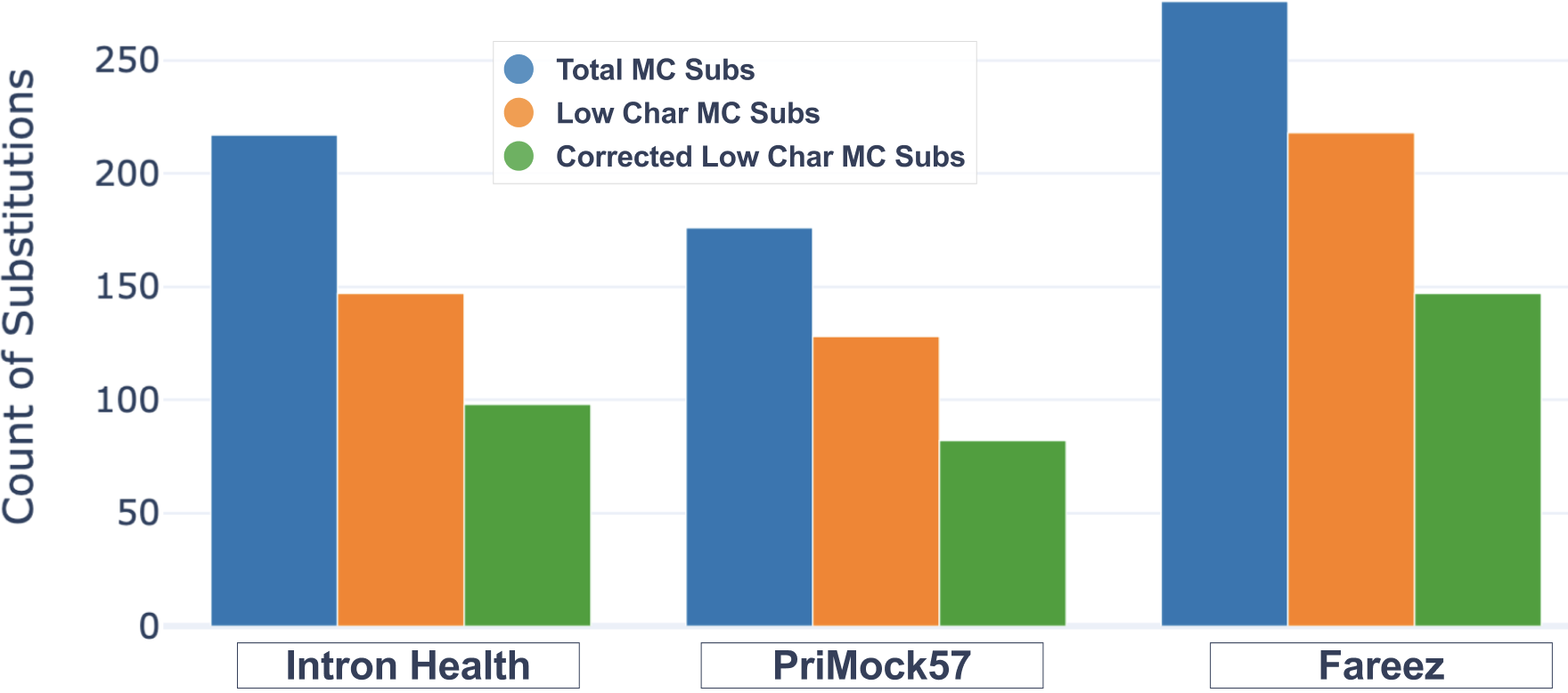}
    \caption{Comparison of total medical concept (MC) substitutions (blue), low character-difference MC substitutions (<5 characters) (orange), and corrected low character-difference MC substitutions (green) across datasets with NVIDIA Canary-1B ASR with Gemini 1.5 Pro correction.}
    \label{fig:nvidia-gemini-char-error-analysis}
\end{figure}

Analysis of correction patterns, using NVIDIA Canary-1B with Gemini 1.5 Pro as a representative model pairing, reveals that effectiveness varies significantly between different types of medical concepts and error categories (Figure \ref{fig:nvidia-gemini-char-error-distribution}). In terms of substitutions, the most substantial improvements were observed in medical problems (conditions and diseases; -63 substitutions), medications (therapeutic drugs and preparations; -44 substitutions), and laboratory data (bodily sample test results; -26 substitutions). In terms of insertions, LLM correction was particularly effective in reducing medication route errors (medication administration locations; -54 insertions) and medical problems (-42 insertions). However, we observed some categories where LLM correction introduced additional errors, notably in medications (+12 insertions), substance abuse terminology (psychoactive substance use; +10 insertions), and biological measurement results (basic vital signs and measurements; +6 insertions). For deletions, correction showed improvements in medical problems (-21 deletions), procedures (diagnostics or treatments; -15 deletions), laboratory data (-10 deletions), and body functions (-10 deletions). The correction process also introduced omissions, particularly for substance abuse terminology (+6 deletions) and body measurement values (+4 deletions).

Analysis of character-level error patterns (Figure \ref{fig:nvidia-gemini-char-error-analysis}) reveals that most medical concept substitution errors involved minor variations of less than 5 characters, comprising 68\% in Intron Health, 73\% in PriMock57, and 79\% in Fareez. LLM correction successfully resolved approximately two thirds of these low character difference errors in all datasets (67\%, 64\%, and 67\%, respectively), demonstrating particular effectiveness in addressing orthographic variations in medical terminology.

\section{Discussion}

\subsection{Impact of Accents on Medical ASR}
Our results demonstrate significant accent-based disparities in ASR performance across medical contexts. The systematic differences between datasets likely reflect the ASR systems' training data composition, with performance declining for non-US accents. This training bias becomes particularly pronounced for non-Western accents, as shown by the substantially higher error rates in Nigerian-accented speech. Interestingly, while accented speech might be expected to disproportionately impact medical terminology recognition, our MC-WER analysis reveals a more nuanced pattern.

\subsection{Comparing WER and MC-WER} 
Our analysis revealed an unexpected pattern: MC-WER values in the PriMock57 and Fareez datasets were consistently lower than their corresponding WER values. This apparent contradiction of medical terms being easier to transcribe requires careful interpretation. MC-WER and WER operate on fundamentally different sampling distributions: while WER considers all words independently, MC-WER focuses exclusively on a subset of medically relevant terms. This means that direct numerical comparisons between WER and MC-WER should be interpreted with caution.

Consider how MC-WER handles compound errors: when "hypertension" becomes "high tension", traditional WER counts both substitution and insertion errors, while MC-WER records only a single substitution. More critically, in a substitution where "amoxicillin" becomes "ampicillin," MC-WER treats this as a simple substitution, despite the significant clinical implications of confusing these antibiotics, which differ in their spectrum of activity and clinical indications. This suggests that while MC-WER effectively captures the preservation of medical terms as linguistic units, it may not reflect the clinical impact of transcription errors.

These findings highlight the need for additional metrics that can bridge the gap between statistical error rates and clinical significance.

\subsection{Comparing LLM and ASR Diarization}
Our results indicate that LLMs consistently diarize at or above the performance of ASR systems in all datasets tested. When examining speaker-specific error patterns, we observed consistently lower error rates in doctor speech compared to patient speech across all datasets. This pattern aligns with the linguistic characteristics of medical conversations: doctors typically follow more standardized patterns of medical discourse, while patients exhibit greater variability in accent and expression. This disparity is most pronounced in the Intron Health dataset, where both doctors and patients speak Nigerian-accented English, but patients' more varied and colloquial expression patterns led to higher error rates. A similar pattern appears in the PriMock57 dataset, where patient speech exhibits various European and non-English accents, while physicians maintain more consistent British English medical discourse, highlighting the ongoing challenges ASR systems face with accent diversity.

\subsection{Differential Impact of LLM Correction}

The effectiveness of LLM correction showed a distinct relationship with baseline ASR performance. Lower-performing ASR systems like Azure STT and NVIDIA Canary-1B saw substantial WER reductions with LLM correction, while high-performing ASR systems like Gemini 1.5 Pro and Soniox showed minimal gains or degradation. These results suggest that high-performing ASR systems may be approaching the theoretical performance limits on these datasets. Interestingly, while Gemini 1.5 Pro's attempts to correct its own transcriptions consistently led to degraded performance, Soniox showed modest improvements when corrected by Gemini 1.5 Pro. This contrast suggests that combining models with complementary strengths—potentially arising from different training focuses, data sources, or optimization objectives—can lead to better overall performance, aligning with ensemble and mixture-of-experts principles in machine learning.

The success of LLM correction was strongly correlated with the type of transcription error. LLMs excelled at correcting minor character-level variations (e.g., "fexifenadine" to "fexofenadine"), but struggled with semantically plausible alternatives (e.g., "relieved" to "really eased") and more distant medical term substitutions (e.g., "metformin" to "warfarin"). These findings suggest that integrated ASR-LLM systems might benefit from specialized pathways for handling semantic, distant medical term substitutions, and orthographic errors.

\section{Conclusion}\label{sec:conclusion}
This study presents a large-scale evaluation of ASR performance and LLM-based corrections in healthcare settings in Nigeria, the United Kingdom, and the United States. Our findings demonstrate that LLMs diarize better than ASR systems in challenging accent scenarios, while their effectiveness in error correction varies with baseline ASR performance. Our analysis of accents and medical terminology highlights the limitations of current metrics such as WER and MC-WER, emphasizing the need for metrics that better reflect the clinical impact of transcription errors.

By evaluating off-the-shelf ASR systems and LLMs without specialized fine-tuning, our findings demonstrate both the potential and current limitations of these tools in improving medical transcription accuracy in global settings. This is particularly relevant for resource-limited environments where access to customized solutions may be impractical. While LLMs show promise in improving medical transcription accuracy, their effectiveness varies by accent and context. Future work should focus on developing more robust solutions that better handle accents and preserve medical terms, particularly for accents underrepresented in current ASR training data.

\section{Limitations}\label{sec:limitations}
Several limitations should be acknowledged in our study. Our focus on English-language medical conversations excludes local languages and code-switching scenarios common in multilingual healthcare settings. The simulated nature of our datasets, while allowing controlled comparison, may not fully capture real-world clinical environments with background noise and varied acoustic conditions.

Our evaluation of LLM correction focused on text-based corrections without taking advantage of acoustic features of the original audio, potentially overlooking valuable phonetic cues for accent-specific error correction. Additionally, while MC-WER quantifies errors in medical terminology, it does not capture the clinical significance of transcription errors, treating all substitutions equally regardless of their potential impact on patient care.

\bibliography{custom}

\appendix

\section{Appendix}
\label{sec:appendix-methodology}

\subsection{Speech Recognition Models}
\label{subsec:appendix-asr-models}
In our analysis, we employed six ASR systems, selected for their advanced capabilities in processing complex medical language and their potential to handle diverse accents and linguistic variations. These systems are:

\begin{enumerate}
  \item Google's Gemini-1.5-Pro (Version 001): Google's latest advancement in multimodal AI as of July 2024, capable of processing both textual and audio inputs simultaneously \citep{GoogleCloudGemini}.
  \item Microsoft Azure's Speech-to-Text: a commercially available service with support for multiple languages and specialized medical vocabulary \citep{AzureAISpeech}.
  \item OpenAI's Whisper 3: the latest iteration of OpenAI's open-source speech recognition model, known for its multilingual capabilities \citep{OpenAIWhisper3}.
  \item NVIDIA NeMo Canary-1B: a multi-lingual, multi-tasking model supporting ASR in English, German, French, and Spanish, with 1 billion parameters \citep{NVIDIANeMOCanary1B}.
  \item Amazon Transcribe Medical: a commercially available service specialized for healthcare. We used the conversation model with specialty set to Primary Care \citep{AWSTranscribeMedical}.
  \item Soniox: a commercially available service. We used the en\_v2 model, which offers high accuracy across various accents \citep{Soniox}.
\end{enumerate}

\subsection{Large Language Models}
\label{subsec:appendix-llms}
To investigate the potential of Large Language Models (LLMs) in improving medical ASR outputs, we selected three models:

\begin{enumerate}
  \item Google's Gemini-1.5-Pro (Version 001): Google's latest advancement in multimodal AI as of July 2024, capable of processing both textual and audio inputs simultaneously \citep{GoogleCloudGemini}.
  \item Anthropic's Claude 3.5 Sonnet: Known for its strong performance in reasoning tasks, it is well suited to analyze and refine medical transcriptions \citep{AnthropicClaude3.5Sonnet}.
  \item OpenAI's GPT-4o: A leading model in natural language processing, offering advanced capabilities in understanding context and nuance in medical conversations \citep{OpenAIGPT4}.
\end{enumerate}

These models were chosen based on their demonstrated capabilities in understanding and generating human-like text, their ability to process and reason over complex information, and their potential for zero-shot and few-shot learning in specialized domains \citep{kojima2023largelanguagemodelszeroshot}.

\subsection{Healthcare Natural Language API}
\label{subsec:appendix-medical-concepts}
To accurately identify and categorize medical concepts within our transcriptions, we employed Google Cloud's Healthcare Natural Language API. This choice builds upon our previous work \cite{adedeji2024sound}, where we demonstrated the API's effectiveness in parsing and structuring unstructured medical text. The Healthcare Natural Language API offers several key functionalities crucial to our study:

\begin{enumerate}
  \item Extraction of medical concepts, including diseases, medications, and procedures.
  \item Categorization of these concepts into over thirty distinct entities.
  \item Analysis of functional features such as temporal relationships and certainty assessments.
\end{enumerate}

A significant advantage of this API is its ability to map extracted concepts to standard medical vocabularies like RxNorm, ICD-10, and SNOMED CT. This mapping is essential for maintaining consistency and accuracy in our medical concept identification across multi-regional datasets.

\subsection{Example Punctuation Prompt}
\label{sec:appendix-example-punctuation-prompt}
For Punctuation, the following prompt template was utilized:\\
\begin{codeblock}
You are a helpful speech-to-text transcription assistant. Your task is to correct punctuation in medical dialogue transcripts, ensuring accurate reflection of natural pauses and speaker transitions. Here's how to approach the task step by step:\\

\noindent1. Contextual Reading: Analyze each sentence for natural pauses, transitions, and speaker changes.\\
\noindent2. Sentence Splitting: Split run-on sentences into separate statements when there's a change in speaker.\\
\noindent3. Question Identification: Mark questions appropriately with question marks, paying attention to intonation and structure.\\
\noindent4. Run-On Correction: Divide run-on sentences into properly punctuated independent clauses.\\
\noindent5. Speaker Transition: Use appropriate punctuation to separate clauses spoken by different speakers.\\
\noindent6. Capitalization: Begin each new statement or speaker transition with a capital letter.\\
\noindent7. Response Delineation: End each response with appropriate terminal punctuation.\\

\noindent8. Examples \string#1-10:\\
\{examples\}\\

\noindent Transcript:\\
\{transcript\_segments\}\\

\noindent Expected output structure as a JSON array:
\begin{verbatim}
[
 {
   "sentence": "[punctuated sentence]"
 },
 {
   "sentence": "[punctuated sentence]"
 }
]
\end{verbatim}
\end{codeblock}

\subsection{Example Diarization Prompt}
\label{sec:appendix-example-diarization-prompt}

For Diarization, the following prompt template was utilized:\\

\begin{codeblock}
You are a helpful speech-to-text transcription assistant. Your current task is to diarize a conversation with no speaker labels. You will use your advanced understanding of medical terminology, dialogue structure, and conversational context to diarize the text accurately. Here's how to approach the task step by step:\\
1. Contextual Reading: Read each sentence thoroughly, absorbing its content, tone, sentiment, and vocabulary.\\
2. Sentence Splitting: Actively split sentences into separate statements when there's a change in speaker. Look for cues like pauses, speech direction changes, thought conclusions, questions, and answers.\\
3. Reasoning: Consider whether the language is technical (suggesting a medical professional) or expresses personal experiences/emotions (suggesting the patient).\\
4. Look-Around Strategy: Analyze the five sentences before and after the current one to understand the conversation flow. Questions may be followed by answers, and concerns by reassurance.\\
5. Label with Justification: Assign a label 'Doctor' or 'Patient' to each sentence, providing a brief justification based on your analysis. Ensure each justification pertains to only one person.\\
6. Consistent Attribution: Maintain a thorough approach throughout the transcript, treating each sentence with equal attention to detail.\\
7. Extremely Granular Attribution: Break down the conversation into the smallest parts (question, answer, utterance) for clarity. Each clause should be precisely attributed to either the doctor or the patient, with no overlap in speaker identity.\\
8. Examples \string#1-10:\\
\{examples\}\\

\noindent Transcript:\\
\{transcript\_segments\}\\

\noindent Expected output structure as a JSON array:
\begin{verbatim}
[
 {
   "sentence": "[reference]",
   "justification": "[justification]",
   "speaker": "Patient or Doctor"
 },
 {
   "sentence": "[reference]",
   "justification": "[justification]", 
   "speaker": "Patient or Doctor"
 }
]
\end{verbatim}
\end{codeblock}

\subsection{Example Correction Prompt}
\label{sec:appendix-example-correction-prompt}
For Correction, the following prompt template was utilized:\
\begin{codeblock}
You are a helpful speech-to-text transcription assistant. Your task is to review and correct transcription errors in medical dialogues, focusing on accuracy and medical context. Here's how to approach the task step by step:\\

\noindent1. Contextual Reading: Analyze each sentence for potential transcription errors, considering medical terminology and context.

\noindent2. Medical Term Verification: Pay special attention to medical terms, medication names, and procedures that may have been misinterpreted.

\noindent3. Accent Consideration: Account for diverse accents and potential misinterpretations in the transcription.

\noindent4. Homophone Analysis: Identify and correct instances where similar-sounding words may have been confused.

\noindent5. Semantic Preservation: Ensure corrections maintain the original meaning while improving accuracy.

\noindent6. Contextual Coherence: Verify that corrected terms align with the medical context of the conversation.\

\noindent7. Examples \string#1-10:\\
\{examples\}\\

\noindent Transcript:\\
\{transcript\_segments\}\
\\[1em]
\noindent Expected output structure as a JSON array:
\begin{verbatim}
[
 {
   "reference_sentence": "[original]",
   "rationale": "[brief explanation]",
   "corrected_sentence": "[corrected]",
   "speaker": "Patient or Doctor"
 },
 {
   "reference_sentence": "[original]",
   "rationale": "[brief explanation]",
   "corrected_sentence": "[corrected]",
   "speaker": "Patient or Doctor"
 }
]
\end{verbatim}
\end{codeblock}

\clearpage
\onecolumn
\subsection{Intron Health WER Table}
\label{subsec:intron health wer table}
\begin{table}[H]
\centering
\fontsize{10pt}{11pt}\selectfont
\setlength{\tabcolsep}{12pt}
\begin{tabular}{| l l l r |}
\hline
LLM & STT & Method & WER \\
\hline
-- & Gemini 1.5 Pro & ASR & 10.18\% $\pm$ 1.81 \\
Gemini 1.5 Pro & Gemini 1.5 Pro & Corrected & 12.86\% $\pm$ 2.09 \\
-- & Amazon Transcribe Medical & ASR & 14.21\% $\pm$ 2.89 \\
Claude 3.5 Sonnet & Gemini 1.5 Pro & Diarized & 15.32\% $\pm$ 6.14 \\
GPT-4o & Soniox & Corrected & 16.28\% $\pm$ 2.72 \\
Gemini 1.5 Pro & Soniox & Corrected & 16.33\% $\pm$ 3.90 \\
-- & Soniox & ASR & 17.54\% $\pm$ 7.89 \\
Gemini 1.5 Pro & Amazon Transcribe Medical & Corrected & 18.24\% $\pm$ 3.30 \\
Claude 3.5 Sonnet & Whisper Large 3 & Corrected & 18.34\% $\pm$ 2.90 \\
Gemini 1.5 Pro & Whisper Large 3 & Corrected & 18.35\% $\pm$ 4.24 \\
GPT-4o & Amazon Transcribe Medical & Corrected & 19.07\% $\pm$ 3.24 \\
-- & Whisper Large 3 & ASR & 19.33\% $\pm$ 2.58 \\
Gemini 1.5 Pro & NVIDIA Canary 1b & Corrected & 21.03\% $\pm$ 2.60 \\
GPT-4o & NVIDIA Canary 1b & Corrected & 21.22\% $\pm$ 2.75 \\
Gemini 1.5 Pro & Azure STT & Corrected & 23.90\% $\pm$ 2.13 \\
GPT-4o & Azure STT & Corrected & 24.11\% $\pm$ 0.00 \\
-- & NVIDIA Canary 1b & ASR & 24.16\% $\pm$ 2.07 \\
-- & Azure STT & ASR & 27.42\% $\pm$ 5.69 \\
\hline
\end{tabular}
\caption{Comparison of WER ASR baselines and their top two LLM corrections in the Intron Health dataset.}
\label{tab:intron_health_wer_results}
\end{table}

\subsection{PriMock57 WER Table}
\label{subsec:primock57 wer table}
\begin{table}[H]
\centering
\fontsize{10pt}{11pt}\selectfont
\setlength{\tabcolsep}{12pt}
\begin{tabular}{| l l l r |}
\hline
LLM & STT & Method & WER \\
\hline
-- & Gemini 1.5 Pro & ASR & 9.34\% $\pm$ 2.69 \\
-- & Amazon Transcribe Medical & ASR & 9.98\% $\pm$ 2.13 \\
Gemini 1.5 Pro & Gemini 1.5 Pro & Corrected & 10.43\% $\pm$ 3.10 \\
Gemini 1.5 Pro & Soniox & Corrected & 11.74\% $\pm$ 3.17 \\
-- & Soniox & ASR & 12.09\% $\pm$ 4.95 \\
Gemini 1.5 Pro & Amazon Transcribe Medical & Corrected & 12.10\% $\pm$ 2.96 \\
Gemini 1.5 Pro & Azure STT & Corrected & 12.19\% $\pm$ 2.58 \\
GPT-4o & Gemini 1.5 Pro & Corrected & 12.43\% $\pm$ 4.88 \\
GPT-4o & Soniox & Corrected & 12.49\% $\pm$ 3.76 \\
GPT-4o & Amazon Transcribe Medical & Corrected & 12.73\% $\pm$ 2.48 \\
GPT-4o & Whisper Large 3 & Corrected & 13.60\% $\pm$ 2.16 \\
Claude 3.5 Sonnet & Whisper Large 3 & Corrected & 13.74\% $\pm$ 2.95 \\
GPT-4o & Azure STT & Corrected & 14.56\% $\pm$ 3.05 \\
Gemini 1.5 Pro & NVIDIA Canary 1b & Corrected & 15.32\% $\pm$ 4.23 \\
-- & Whisper Large 3 & ASR & 16.76\% $\pm$ 2.52 \\
Claude 3.5 Sonnet & NVIDIA Canary 1b & Corrected & 17.35\% $\pm$ 4.55 \\
-- & Azure STT & ASR & 18.43\% $\pm$ 2.67 \\
-- & NVIDIA Canary 1b & ASR & 20.65\% $\pm$ 2.61 \\
\hline
\end{tabular}
\caption{Comparison of WER ASR baselines and their top two LLM corrections in the PriMock57 dataset.}
\label{tab:primock57_wer_results}
\end{table}

\clearpage
\subsection{Fareez WER Table}
\label{subsec:fareez wer table}
\begin{table}[H]
\centering
\fontsize{10pt}{11pt}\selectfont
\setlength{\tabcolsep}{12pt}
\begin{tabular}{| l l l r |}
\hline
LLM & STT & Method & WER \\
\hline
-- & Gemini 1.5 Pro & ASR & 6.19\% $\pm$ 2.30 \\
-- & Amazon Transcribe Medical & ASR & 6.33\% $\pm$ 1.89 \\
Gemini 1.5 Pro & Amazon Transcribe Medical & Corrected & 7.42\% $\pm$ 2.43 \\
-- & Soniox & ASR & 8.14\% $\pm$ 3.33 \\
Gemini 1.5 Pro & Soniox & Corrected & 8.28\% $\pm$ 1.98 \\
Gemini 1.5 Pro & Azure STT & Corrected & 8.45\% $\pm$ 2.15 \\
Claude 3.5 Sonnet & Gemini 1.5 Pro & Diarized & 8.73\% $\pm$ 10.72 \\
Claude 3.5 Sonnet & Amazon Transcribe Medical & Diarized & 9.05\% $\pm$ 3.33 \\
Claude 3.5 Sonnet & Soniox & Corrected & 9.15\% $\pm$ 2.23 \\
Gemini 1.5 Pro & NVIDIA Canary 1b & Corrected & 9.78\% $\pm$ 2.19 \\
Gemini 1.5 Pro & Gemini 1.5 Pro & Corrected & 9.79\% $\pm$ 9.11 \\
Claude 3.5 Sonnet & Whisper Large 3 & Diarized & 10.34\% $\pm$ 5.37 \\
Claude 3.5 Sonnet & Azure STT & Diarized & 11.08\% $\pm$ 6.28 \\
Claude 3.5 Sonnet & NVIDIA Canary 1b & Diarized & 11.49\% $\pm$ 5.33 \\
Claude 3.5 Sonnet & Whisper Large 3 & Corrected & 11.95\% $\pm$ 2.47 \\
-- & Azure STT & ASR & 14.29\% $\pm$ 1.98 \\
-- & Whisper Large 3 & ASR & 14.30\% $\pm$ 2.79 \\
-- & NVIDIA Canary 1b & ASR & 14.89\% $\pm$ 2.64 \\
\hline
\end{tabular}
\caption{Comparison of WER ASR baselines and their top two LLM corrections in the Fareez dataset.}
\label{tab:fareez_wer_results}
\end{table}

\subsection{Intron Health MC-WER Table}
\label{subsec:intron health mc-wer table}
\begin{table}[H]
\centering
\fontsize{10pt}{11pt}\selectfont
\setlength{\tabcolsep}{12pt}
\begin{tabular}{| l l l r |}
\hline
LLM & STT & Method & WER \\
\hline
-- & Gemini 1.5 Pro & ASR & 10.69\% $\pm$ 3.72 \\
Gemini 1.5 Pro & Gemini 1.5 Pro & Corrected & 12.54\% $\pm$ 4.38 \\
GPT-4o & Soniox & Diarized & 16.83\% $\pm$ 5.78 \\
Claude 3.5 Sonnet & Soniox & Diarized & 17.14\% $\pm$ 6.79 \\
Claude 3.5 Sonnet & Gemini 1.5 Pro & Corrected & 17.45\% $\pm$ 6.27 \\
-- & Soniox & ASR & 18.12\% $\pm$ 7.60 \\
-- & Whisper Large 3 & ASR & 19.46\% $\pm$ 6.74 \\
Claude 3.5 Sonnet & Whisper Large 3 & Corrected & 19.95\% $\pm$ 7.73 \\
GPT-4o & Whisper Large 3 & Corrected & 20.34\% $\pm$ 6.89 \\
-- & Amazon Transcribe Medical & ASR & 20.89\% $\pm$ 8.19 \\
Gemini 1.5 Pro & Amazon Transcribe Medical & Corrected & 21.16\% $\pm$ 8.58 \\
GPT-4o & Amazon Transcribe Medical & Corrected & 21.66\% $\pm$ 7.11 \\
Claude 3.5 Sonnet & NVIDIA Canary 1b & Corrected & 25.52\% $\pm$ 6.67 \\
Gemini 1.5 Pro & NVIDIA Canary 1b & Corrected & 25.76\% $\pm$ 6.69 \\
-- & NVIDIA Canary 1b & ASR & 30.16\% $\pm$ 8.62 \\
GPT-4o & Azure STT & Corrected & 46.17\% $\pm$ 11.86 \\
Gemini 1.5 Pro & Azure STT & Corrected & 47.82\% $\pm$ 12.27 \\
-- & Azure STT & ASR & 48.67\% $\pm$ 12.60 \\
\hline
\end{tabular}
\caption{Comparison of MC-WER ASR baselines and their top two LLM corrections in the Intron Health dataset.}
\label{tab:intron_health_mc_wer_results}
\end{table}

\clearpage
\subsection{PriMock57 MC-WER Table}
\label{subsec:primock57 mc-wer table}
\begin{table}[H]
\centering
\fontsize{10pt}{11pt}\selectfont
\setlength{\tabcolsep}{12pt}
\begin{tabular}{| l l l r |}
\hline
LLM & STT & Method & WER \\
\hline
-- & Gemini 1.5 Pro & ASR & 7.38\% $\pm$ 4.17 \\
Gemini 1.5 Pro & Gemini 1.5 Pro & Corrected & 8.13\% $\pm$ 4.60 \\
Gemini 1.5 Pro & Soniox & Corrected & 8.35\% $\pm$ 4.25 \\
-- & Soniox & ASR & 8.38\% $\pm$ 4.30 \\
Claude 3.5 Sonnet & Soniox & Diarized & 8.84\% $\pm$ 4.34 \\
GPT-4o & Gemini 1.5 Pro & Corrected & 9.26\% $\pm$ 5.29 \\
-- & Amazon Transcribe Medical & ASR & 9.47\% $\pm$ 4.23 \\
-- & Whisper Large 3 & ASR & 9.57\% $\pm$ 4.51 \\
Gemini 1.5 Pro & Amazon Transcribe Medical & Corrected & 10.40\% $\pm$ 4.93 \\
Claude 3.5 Sonnet & Whisper Large 3 & Corrected & 10.63\% $\pm$ 5.24 \\
GPT-4o & Whisper Large 3 & Corrected & 10.82\% $\pm$ 5.49 \\
Gemini 1.5 Pro & Azure STT & Corrected & 11.62\% $\pm$ 4.90 \\
-- & Azure STT & ASR & 11.76\% $\pm$ 5.06 \\
GPT-4o & Amazon Transcribe Medical & Corrected & 11.82\% $\pm$ 5.58 \\
Gemini 1.5 Pro & NVIDIA Canary 1b & Corrected & 13.25\% $\pm$ 5.97 \\
-- & NVIDIA Canary 1b & ASR & 13.26\% $\pm$ 5.84 \\
Claude 3.5 Sonnet & Azure STT & Corrected & 13.68\% $\pm$ 6.60 \\
Claude 3.5 Sonnet & NVIDIA Canary 1b & Corrected & 15.22\% $\pm$ 6.98 \\
\hline
\end{tabular}
\caption{Comparison of MC-WER ASR baselines and their top two LLM corrections in the PriMock57 dataset.}
\label{tab:primock57_mc_wer_results}
\end{table}

\subsection{Fareez MC-WER Table}
\label{subsec:fareez mc-wer table}
\begin{table}[H]
\centering
\fontsize{10pt}{11pt}\selectfont
\setlength{\tabcolsep}{12pt}
\begin{tabular}{| l l l r |}
\hline
LLM & STT & Method & WER \\
\hline
-- & Soniox & ASR & 3.28\% $\pm$ 3.02 \\
Claude 3.5 Sonnet & Soniox & Diarized & 3.47\% $\pm$ 3.20 \\
-- & Gemini 1.5 Pro & ASR & 3.58\% $\pm$ 3.47 \\
Gemini 1.5 Pro & Soniox & Corrected & 3.81\% $\pm$ 3.58 \\
-- & Azure STT & ASR & 4.22\% $\pm$ 3.01 \\
Claude 3.5 Sonnet & Gemini 1.5 Pro & Corrected & 4.37\% $\pm$ 3.40 \\
-- & Amazon Transcribe Medical & ASR & 4.41\% $\pm$ 3.45 \\
Gemini 1.5 Pro & Amazon Transcribe Medical & Corrected & 4.58\% $\pm$ 4.11 \\
-- & Whisper Large 3 & ASR & 5.00\% $\pm$ 3.63 \\
Gemini 1.5 Pro & Azure STT & Corrected & 5.07\% $\pm$ 3.49 \\
Claude 3.5 Sonnet & Amazon Transcribe Medical & Corrected & 5.16\% $\pm$ 3.85 \\
Claude 3.5 Sonnet & Azure STT & Corrected & 6.53\% $\pm$ 3.96 \\
Gemini 1.5 Pro & NVIDIA Canary 1b & Corrected & 6.84\% $\pm$ 4.30 \\
Gemini 1.5 Pro & Whisper Large 3 & Corrected & 6.94\% $\pm$ 4.30 \\
Claude 3.5 Sonnet & Gemini 1.5 Pro & Diarized & 7.07\% $\pm$ 9.35 \\
Claude 3.5 Sonnet & Whisper Large 3 & Corrected & 7.25\% $\pm$ 4.62 \\
-- & NVIDIA Canary 1b & ASR & 7.35\% $\pm$ 4.05 \\
Claude 3.5 Sonnet & NVIDIA Canary 1b & Corrected & 7.62\% $\pm$ 4.25 \\
\hline
\end{tabular}
\caption{Comparison of MC-WER ASR baselines and their top two LLM corrections in the Fareez dataset.}
\label{tab:fareez_mc_wer_results}
\end{table}

\twocolumn
\end{document}